\documentclass[sigconf]{acmart}
\usepackage{amsmath,amsfonts}
\usepackage{mathtools}
\usepackage{epsfig}
\usepackage[mathscr]{euscript}
\usepackage{color}
\usepackage{booktabs} 
\usepackage{stmaryrd} 
\usepackage{bbm} 
\usepackage{blindtext, comment}
\usepackage{rotating} 
\usepackage{tikz}
\usepackage{standalone}
\usetikzlibrary{arrows, patterns}
\usetikzlibrary{decorations.pathreplacing,angles,quotes}
\usepackage{framed, standalone}
\usepackage{graphicx}
\usepackage{capt-of}
\usepackage{balance}
\usepackage{makecell} 
\usepackage{afterpage}
\usepackage[linesnumbered,lined,boxed,commentsnumbered,ruled,vlined]{algorithm2e}
\usepackage{setspace}
\usepackage{listings}
\definecolor{Code}{rgb}{0,0,0}
\definecolor{Decorators}{rgb}{0.5,0.5,0.5}
\definecolor{Numbers}{rgb}{0.5,0,0}
\definecolor{MatchingBrackets}{rgb}{0.25,0.5,0.5}
\definecolor{Keywords}{rgb}{0,0,1}
\definecolor{self}{rgb}{0,0,0}
\definecolor{Strings}{rgb}{0,0.63,0}
\definecolor{Comments}{rgb}{0,0.63,1}
\definecolor{Backquotes}{rgb}{0,0,0}
\definecolor{Classname}{rgb}{0,0,0}
\definecolor{FunctionName}{rgb}{0,0,0}
\definecolor{Operators}{rgb}{0,0,0}
\definecolor{Background}{rgb}{0.98,0.98,0.98}
\usepackage{textcomp}
\usepackage{xcolor}
\def\BibTeX{{\rm B\kern-.05em{\sc i\kern-.025em b}\kern-.08em
    T\kern-.1667em\lower.7ex\hbox{E}\kern-.125emX}}
\AtBeginDocument{%
  \providecommand\BibTeX{{%
    \normalfont B\kern-0.5em{\scshape i\kern-0.25em b}\kern-0.8em\TeX}}}

\setcopyright{acmcopyright}
\copyrightyear{2021}
\acmYear{2021}
\acmDOI{10.1145/1122445.1122456}

\acmConference[KDD '21]{KDD Workshop on Machine Learning in Finance}{August 15, 2021}{Singapore}
\acmBooktitle{KDD Workshop on Machine Learning in Finance, August 15, 2021, Singapore}
\acmPrice{15.00}
\acmISBN{978-1-4503-XXXX-X/18/06}

\begin{document}

\title{XtracTree: a Simple and Effective Method\\for Regulator Validation of Bagging Methods\\Used in Retail Banking}

\author{Jeremy Charlier}
\email{name.surname@bnc.ca}
\affiliation{%
  \institution{National Bank of Canada}
  \city{Montreal}
  \state{Quebec}
  \country{Canada}
}

\author{Vladimir Makarenkov}
\email{surname.name@uqam.ca}
\affiliation{%
    \institution{Université du Québec à Montréal (UQAM)}
  \city{Montreal}
  \state{Quebec}
  \country{Canada}}

\renewcommand{\shortauthors}{Charlier and Makarenkov}

\begin{abstract}
  Bootstrap aggregation, known as bagging, is one of the most popular ensemble methods used in machine learning (ML). An ensemble method is a supervised ML method that combines multiple hypotheses to form a single hypothesis used for prediction. A bagging algorithm combines multiple classifiers modelled on different sub-samples of the same data set to build one large classifier. Large retail banks are nowadays using the power of ML algorithms, including decision trees and random forests, to optimize retail banking activities. However, AI bank researchers face a strong challenges. It starts with their own model validation department, followed by the deployment of the solution in an production environment up to the external validation of the national financial regulators. Each proposed ML model has to be validated and clear rules for every algorithm-based decision have to be established. In this context, we propose XtracTree, an algorithm that is capable of effectively converting an ML bagging classifier, such as a decision tree or a random forest, into simple ``if-then" rules satisfying the requirements of model validation.
  We use a public loan data set from Kaggle to illustrate the usefulness of our approach. Our experiments indicate that, using XtracTree, we are able to ensure a better understanding for our model, leading to an easier model validation by national financial regulators and the internal model validation department. The proposed approach allowed our banking institution to reduce up to 50\% the man-days required to deliver AI solutions into production.
\end{abstract}




\maketitle
\section{Motivation} \label{sec::intro}
\begin{figure*}[t]
    \centering
    \includegraphics[scale=.53]{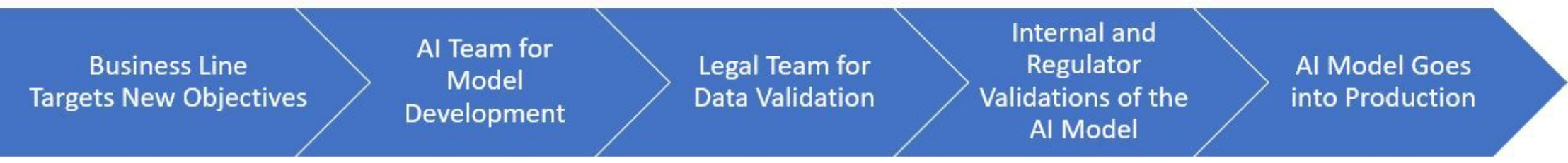}
  \caption{Validation process of a Machine Learning (ML) approach in a retail bank. The approach starts from the need of the business line to achieve their objectives. The AI team defines with them the objectives, the data to be used, and the expected deliverables. The AI team then consults the legal team for a legal approval of the data to be used. Once the AI team finishes its work, the model and the approach need to be validated by the internal validation department. Then, the validation department transfer the details of the work for the regulator approval. The ML approach becomes available to the business line after the approval of the regulator.}
  \label{fig::validationprocess}
\end{figure*}

The digital transformation of developed societies has forced the banks to embrace the power of data and artificial intelligence \cite{king2018bank}. Digital banks in China, for instance, are becoming the most popular banks \cite{Wang2019tech} and have the fastest growth on the market. The clients can avoid queuing in agencies, while having access to instantaneous transactions using a digital platform. This new type of client habits towards the banking system is changing drastically the relationship between the banks and their clients. The banks have to use digital data at their disposal to get more information about their clients. They can no longer rely on the visits to the bank agencies. The banks are consequently transitioning from a human-based approach to an advanced data approach with the use of machine learning (ML) techniques \cite{Ng2020hands}. A large variety of ML banking applications focuses on reducing the risk of losses of certain types of retail banking activities such as credit-card limits or online loan applications. The banks have, however, strong constraints regarding the use of the data and ML algorithms. They have mandatory requirements from the internal model validation departments and the regulator validations for all models used in production. We describe in Figure \ref{fig::validationprocess} a standard validation process of a new ML-based implementation in a retail bank. The current challenge is, therefore, to design efficient and explainable algorithms which will be approved at each step of the process by different validation groups. To do so, the models have to extract rules that can be easily understandable by non-ML specialists. This requirement is at the core of our approach.\\

An extensive use of neural networks and other machine learning algorithms over the recent years \cite{bengio2017deep, lecun2015deep, alpaydin2020introduction} has confirmed that some algorithms are more suitable for certain applications than others. On the one hand, the Convolutional Neural Networks (CNNs) \cite{fukushima1980neocognitron,lecun1999object} are usually outperforming all the algorithms for image applications \cite{ren2015exploring,chen2017sca}. On the other hand, the financial data sets mostly contain structured data mixing categorical and continuous variables, which require supervised learning. To the best of our knowledge, no work on the use of bagging methods compliant to financial regulations has been published, mostly due to the confidentiality issues. Well-known in the ML community and the ML competitions\footnote{https://www.kaggle.com/}, the boosting and the bootstrap aggregation (bagging) algorithms constitute the winning choice in about 70\% of ML competitions \cite{chen2016xgboost}. We recall that a gradient boosting algorithm builds iteratively a prediction model based on a set of weak prediction models, called weak learners \cite{friedman2001greedy}. A bagging algorithm builds a collection of independent predictors and then combines them to create a prediction model \cite{breiman1996bagging}. Lately, the boosting algorithms have been preferred to bagging algorithms because of their success in ML competitions \cite{chen2016xgboost}. A large collection of boosting algorithms are consequently available, such as GradientBoosting \cite{friedman2001greedy} or AdaBoost \cite{freund1995desicion,hastie2009multi}, while the most popular among them are XGBoost \cite{chen2016xgboost} and lightGBM \cite{ke2017lightgbm}. XGBoost is a sparsity-aware algorithm for sparse data and 
LightGBM aims at building an efficient model on large data sets. 
Nonetheless, the main challenge for ML algorithms 
applied in a retail bank supervised by a financial regulator is not to provide the best values of evaluation metrics such as the accuracy or F-score. The main challenge here is the explainability of the algorithm results and their cross-validation with the results of other algorithms.
\\

In this work, we aim at building an algorithm that will ease the validation procedure of ML bagging algorithms in retail banking by the financial regulator and the internal model validation  department. Our approach uses the popular Random Forest (RF) algorithm that allows one to build a collection of decision trees to make predictions \cite{breiman1996bagging,breiman2000some,breiman2001random,breiman2004consistency}, as it is often challenging to validate the predictions made by either XGBoost or lightGBM. Our algorithm extracts the decision path of the RF to convert it into a set of ``if-then" rules. These rules are designed to be understandable by a large audience. We show in our experiments that the derived rules can be used to make predictions of the same accuracy as the original classifier, and to draw the decision path for each individual sample. The validation process of ML algorithms used within large banks is at the core of our approach as the AI teams have to focus on the explainability of their model. Our main contributions are summarized below:
\begin{itemize}
    \item We propose a novel algorithm, XtracTree, to convert the decision rules of a RF algorithm into a set of ``if-then" rules. The process of converting the decision rules into ``if-then" rules is crucial to avoid the ``black-box" term usually associated with ML algorithms, especially in the context of model validation by the regulators for large corporation banks.
    \item Our approach uses the set of ``if-then" rules to build accurate predictions, identical to the predictions of the original algorithm. It is crucial for the acceptance of the model by the business line that will later use it.
    \item Our algorithm underlines the decision path of each prediction. It is a strong validation requirement that the model has to demonstrate the features and the decision path that led to a retail banking decision, such as the approval of a new credit card. This is also a requirement from regulations such as General Data Protection Regulation (GDPR).
    \item We demonstrate that efficient ML solutions can be deployed in heavily regulated environment such as the banking sector. It highlights how our algorithm is currently used across different data and ML projects in our banking institution to reduce the time to delivery independently of the production environment chosen.
\end{itemize}

The paper is structured as follows. We discuss the related work in Section \ref{sec::relwork}. We briefly review the random forest approach in Section \ref{sec::background}. We then describe in Section \ref{sec::propmethod} how XtracTree is capable of converting a RF algorithm into simple ``if-then" rules to address the explainability concerns of the validation department and the regulator. We underline that the created ``if-then" rules are still capable of making accurate predictions, while being able to backtrack the decision process for each sample. We thus discuss how Xtractree helped to accelerate the time to delivery of the end product while facing deployment challenges for different production platforms. We describe the experimental results in Section \ref{sec::exp}. Finally, we conclude and address promising directions for future work.
\\

\section{Related Work} \label{sec::relwork}
XtracTree uses the RF algorithms \cite{breiman1996bagging,breiman2000some,breiman2001random,breiman2004consistency} to provide a set of rules which will ease the validation of the model. We recall that RFs use an ensemble of trees, where each tree is grown according to a random parameter \cite{biau2012analysis}. The final model is obtained by aggregating all the trees. The advantages of RFs are numerous. They are capable of producing accurate predictions, while being fast and easy to implement in comparison to the state-of-the-art boosting algorithms \cite{biau2012analysis}. RFs also cope well with the overfitting of large data sets and, therefore, constitute one of the most accurate general purpose learning techniques. RFs, consequently, are in line with our objective of providing explainable rules, leading to accurate predictions, for both the internal and the external model validation requirements of a large retail bank. The success of  RFs is highlighted by a large number of applications described in numerous research publications. Recent publications, such as \cite{kong2018deep} or \cite{sathe2019nearest}, rely on RFs to either extract feature representation in gene expression or to validate new classification algorithms. RFs have been applied successfully in finance as well. In \cite{meng2019practice}, RFs are used to estimate the future risk of default of the borrowers to allow the risk mitigation of the bank's portfolio. In \cite{mercadier2019credit}, RFs are also used for credit-related applications including the credit default swaps spreads, which play a role in the estimation of the default probability. However, to the best of our knowledge, no references are available regarding the use of RFs in retail banking to facilitate the model validation by deriving a simple set of rules targeting a larger audience than pure ML specialists.
\\

RFs and the bagging methods are often concurrent to the boosting methods. The boosting methods were introduced simultaneously with RFs \cite{freund1996experiments}. Boosting trees became extremely popular with XGBoost \cite{chen2016xgboost} and LightGBM \cite{ke2017lightgbm} thanks to different machine learning competitions organized by Kaggle\footnote{https://www.kaggle.com/}. We recall that XGBoost is a sparsity-aware algorithm for sparse data and weighted quantile sketch for approximate tree learning. 
LightGBM combines Gradient-based One-Side Sampling (GOSS) with Exclusive Feature Bundling description to build an efficient model on large data sets. 
The advantage of LightGBM over XGBoost is its optimization for large data sets. 
XGBoost has been extensively used in finance. In \cite{chang2018application}, XGBoost was used to improve the credit risk assessment with the aim of improving the loan decisions of different financial institutions. In \cite{akila2017risk}, XGBoost was used to detect the risk of credit fraud. Similarly, lightGBM has been applied in different financial applications such as the credit predictions of mobile users \cite{guo2019mobile} to be used by digital banks or for cryptocurrency predictions \cite{sun2018novel}. We underline that our approach presented in Section \ref{sec::propmethod} can be easily used with any boosting algorithm, including LightGBM and XGBoost.
\\

\section{Background} \label{sec::background}
In this section, we briefly recall some theoretical foundations of the Random Forest (RF) approach \cite{breiman1996bagging,breiman2000some,breiman2001random,breiman2004consistency}. We follow the notations and the description of \cite{biau2012analysis}. A RF is a collection of randomized base regression trees, such as
$\left\{ r_n(\mathbf{x}, \Theta_m, \mathcal{D}_n), m\geq 1 \right\}$,
where
$\Theta_1, \Theta_2, ...$, are independent and identically distributed outputs of a randomizing variable
$\Theta$, $n$ is the number of training samples and $m$ is the number of random trees. The trees are combined to form the aggregated regression estimate
\begin{equation}
\bar{r}_n (\mathbf{X}, \mathcal{D}_n) = \mathbb{E}_\Theta \left[ r_n(\mathbf{X}, \Theta, \mathcal{D}_n) \right] ,
\end{equation}
where $\mathbb{E}_\Theta$ denotes the expectation with respect to the random parameter,  conditionally on $\mathbf{X}$ and the data set $\mathcal{D}_n$. The randomized variable $\Theta$ allows to compute the successive cuts when building the individual trees, such as the selection of the split coordinate.
\\

In this brief background description, we do not consider bootstrapping and resampling techniques for the ease of explanation. The nodes of the trees are therefore associated with rectangular cells. At each step of the tree construction, the collection of cells associated with the leaves of the trees forms a partition of $[0, 1]^d$. The root of the tree is similarly $[0, 1]^d$. The following splitting procedure is repeated $\left \lceil \log_2 k_n \right \rceil$ times, where $\left \lceil . \right \rceil$ is the ceiling function and $k_n \geq 2$ is a deterministic parameter set by the user. At each node, a coordinate of $\mathbf{X}=(X^{(1)}, ..., X^{(d)})$ is selected, the $j$-th feature having the probability $p_{nj} \in (0,1)$ of being selected. Then, the split is computed at the midpoint of the chosen side. Each randomized tree $r_n(\mathbf{X}, \Theta)$ outputs the average for all $Y_i$ for which the corresponding vector $\mathbf{X}_i$ falls into the same cell, as the random partition of $\mathbb{X}$. Therefore, for $A_n(\mathbf{X}, \Theta)$ and the rectangular cell of the random partition containing $X$, we have:
\begin{equation}
r_n(\mathbf{X}, \Theta) = \dfrac{\sum_{i=1}^{n} Y_i \mathbf{1}_{[\mathbf{X}_i\in A_n(\mathbf{X},\Theta)]}}{\sum_{i=1}^n \mathbf{1}_{[\mathbf{X}_i\in A_n(\mathbf{X},\Theta)]}} \mathbf{1}_{\mathcal{E}_n(\mathbf{X}, \Theta)},
\end{equation}
with
\begin{equation}
\mathcal{E}_n(\mathbf{X},\Theta)= \left[ \sum_{i=1}^n \mathbf{1}_{[X_i\in A_n(\mathbf{X},\Theta)]} \neq 0 \right] .
\end{equation}
By taking the expectation according to $\Theta$, the RF estimate takes the form:
\begin{equation}
\begin{split}
\bar{r}_n(\mathbf{X}, \mathcal{D}_n) = &\: \mathbb{E}_\Theta \left[ r_n(\mathbf{X}, \Theta, \mathcal{D}_n) \right] \\
\bar{r}_n(\mathbf{X}, \mathcal{D}_n) = &\: \mathbb{E}_\Theta \left[ \dfrac{\sum_{i=1}^{n} Y_i \mathbf{1}_{[\mathbf{X}_i\in A_n(\mathbf{X},\Theta)]}}{\sum_{i=1}^n \mathbf{1}_{[\mathbf{X}_i\in A_n(\mathbf{X},\Theta)]}} \mathbf{1}_{\mathcal{E}_n(\mathbf{X}, \Theta)} \right ] .
\end{split}
\end{equation}

\section{Proposed Method: XtracTree for Regulator Validation} \label{sec::propmethod}

We propose XtracTree that is a bagging parser to ``if-then" rules, applicable to any decision tree or RF algorithm. This conversion of a simple set of rules with its large scale of applicability is at the core of the validation approach of ML algorithms by financial regulators. In this section, we first describe how to convert a decision tree or a RF into a set of ``if-then" rules. We then explain how to use the extracted set of rules to get accurate predictions. We finally show how to use the model to display the decision path of each client, bringing an extensive explanation of the model's decisions.

\subsection{XtracTree: a bagging parser to ``if-then" rules}

The idea behind XtracTree is fairly simple and inherited from the recurrent questions of the financial regulators and the model validation department. How to provide a simple explanation of the results of our machine learning models for a large audience not familiar with such concepts? We perform a cross-validation technique with well-known machine learning algorithms such as decision trees, and later RFs. We imitate the behavior of the champion machine learning model with a RF, or a decision tree, to easily extract a set of simple ``if-then" rules. The approach includes four main steps. First, we train a bagging model and then  determine the hyper-parameters providing the best performance. Second, we apply XtracTree with the trained bagging model. XtracTree collects the complete decision path leading to each leaf with all the features and the respective threshold values. Third, XtracTree aggregates and converts the decision path rules of each leaf to successive ``if-then" conditions using the extracted features and the respective threshold values. Fourth, the set of ``if-then" rules are written in an independent csv file. XtracTree consequently maps all the trees collection of the bagging classifier into a Python function and generates an output file that could be then reused independently of any machine learning model, free of any compatibility dependencies. We illustrate the approach in Algorithm \ref{algo::XtracTree}.

\begin{figure*}[t]
    \centering
    \includegraphics[scale=.52]{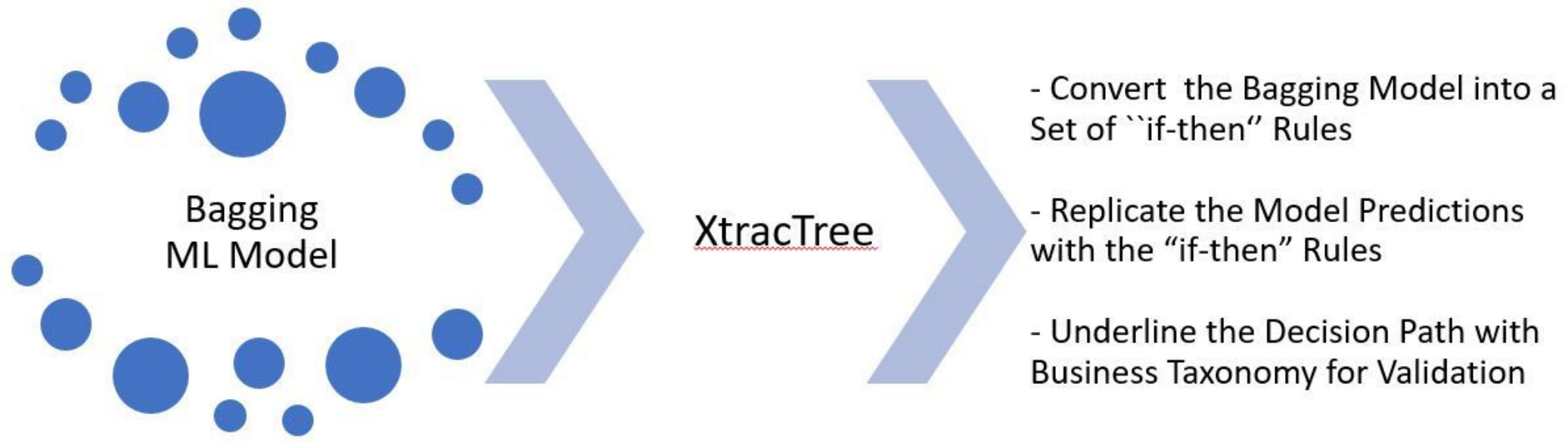}
  \caption{XtracTree highlights the decision path of a bagging machine learning (ML) model. XtracTree builds ``if-then" rules from the bagging model and uses these rules to replicate the model predictions. Due to the ML ``black-box" concern in retail banking, XtracTree uses Business Taxonomy to highlight the decision path for each client.}
  \label{fig::extract_rules}
\end{figure*}

\SetAlFnt{\scriptsize}
\SetAlCapFnt{\scriptsize}
\SetAlCapNameFnt{\scriptsize}

\begin{algorithm}[t!]
\setstretch{1.0}
\DontPrintSemicolon

\KwData{$\mathcal{D}$ data set, trained model}

\KwResult{Output file containing ``if-then" rules}

\Begin{

\textbf{function} \textit{build\_rules($^*$arg)} \{ \\
	$\quad$Retrieve nodes, children\_left, children\_right \\
	$\quad$Retrieve features, thresholds \\
    $\quad$state = 0 \\
	$\quad$\For{each node in nodes}{
		$\quad$\If{node == leaf}{
			$\quad$write(return \%s \% probabilities of occurrences)
		}
		$\quad$\Else{write(state = (if features $<=$ threshold then children\_left else children\_right))}
	}
\}

\textbf{function} \textit{model\_2rules(model)} \{ \\
$\quad$\If{model == RF}{
	$\quad$\For{each Tree in RF}{call build\_rules(current tree)}}
$\quad$\Else{call build\_rules(model)}
\}
}
\KwRet{}

\caption{Extracting Tree Rules for Validation Requirements\label{algo::XtracTree}}
\end{algorithm}

\subsection{Computing Accurate Predictions with Extracted Rules}

The parsing of the bagging model to the ``if-then" rules helps convert a perceived ``black-box" machine learning model by the regulator into a more understandable model for a non-expert. The recurrent question is to how we can be sure that the set of rules extracted by XtracTree are legitimate. As an answer to that question, XtracTree relies on an independent \textit{predict} function. This \textit{predict} function extracts the probabilities of each ``if-then" condition for each leaf of each collection of trees in the bagging model for every sample to predict. The predictor then sums up all the intermediate probabilities per sample to predict the final probability. We refer to Figures \ref{fig::DT} and \ref{fig::RF} in our experiments to present an overview of the ``if-then" rules and the leaf probabilities. The \textit{predict} function consequently builds up on the \textit{model\_2rules} function to obtain a static classifier capable of performing predictions using the ``if-then" rules. We describe in a few lines of codes in Algorithm \ref{algo::predict_XtracTree} how the \textit{predict} function works in relation to the \textit{model\_2rules} function. We underline the accuracy of the approach in our experiments. The static classifier does not, furthermore, impact the model validation as the data sets and the features are fixed and provided to the regulator for the validation procedure.

\SetAlFnt{\scriptsize}
\SetAlCapFnt{\scriptsize}
\SetAlCapNameFnt{\scriptsize}

\begin{algorithm}[t!]
\setstretch{1.0}
\DontPrintSemicolon

\KwData{$\mathcal{D}$ data set, model, XtracTree file containing ``if-then" rules in the function model\_2rules, X\_2predict}

\KwResult{Predicted Probability with ``if-then" Rules}

\Begin{

\textbf{function} \textit{model\_2rules(model)} \{ \\
	$\quad$static ``if-then" rules returning the probability of occurrences \\
\}

\textbf{function} \textit{predict(X\_2predict)} \{ \\
	$\quad$new\_probability = 0
	$\quad$\If{model == RF}{
		$\quad$\For{each Tree in RF}{probability += call model\_2rules (current tree)}}
	$\quad$\Else{probability = call model\_2rules}
	$\quad$return probability\\
\}
}
\KwRet{}

\caption{Predictions using the Extracted Tree Rules for Validation Requirements\label{algo::predict_XtracTree}}
\end{algorithm}

\subsection{Display Any Decision Path On Demand}
In retail banking, all the decisions taken by a model have to be explainable. A client, for instance, can question a process that led to a declined application. We, therefore, put emphasis on developing an approach that allows one to understand the ML model decisions for model validation, and is capable of underlining the decision path for each client. Certain types of financial products are, furthermore, addressed to certain group of clients, such as a complete family. The algorithm, consequently, has to be able to demonstrate the decision path on the complete group. In our approach, we back propagate the extracted ``if-then" decision rules to the fitted bagging model for the samples to be predicted. 
\\

We define a function, called \textit{display\_rule\_per\_estimator}, that will be in charge of highlighting the decision path for either a client or a group of clients. In the case of a client, the \textit{display\_rule\_per\_esti\-mator} function extracts the nodes, the children, the features and the threshold values per estimator; an estimator being one of the collection trees of a RF or a decision tree itself. Then, by comparing the decision rules of the estimator, the function is capable of showing the reasoning behind each prediction in the form of ``if-then" rules. The rules are finally prioritized based on the order of features importance of the model. In the case of a group of clients, the function is identical. It reflects the same reasoning except that it only highlights the common ``if-then" rules among different clients. We finally reach the objective of providing simple rules for model validation as well as providing a human-understandable decision path that leads to the model decision for a client or a group of clients. 

\subsection{Using Xtractree to Help Business Practices in Retail Banking Activities}
The biggest challenge of using data-driven approaches and ML techniques for a retail banking institution is the mandatory compliance to the strong regulations, such as Basel III, the GDPR or the Revised Payment Service Directive (PSD2). All models, including ML algorithms, must be approved by governance and validation, external audit and the financial regulators. Therefore, the time to delivery of any modelling projects impacting the credit risk of a bank may exceed 3 years, from the kick-off origination to the solution deployed into a production environment, such as an AWS container for instance. We designed Xtractree to bypass the fear of the regulators of the ``black-box" ML. Other techniques do exist, such as LIME \cite{ribeiro2016should}, but these techniques are often local approximations and, consequently, do not help to reduce the validation time, nor give more confidence to the regulators about the ML solutions. Thanks to Xtractree, currently used by our development AI teams across different projects, we are able to significantly accelerate the deployment of ML solutions and data-driven approaches because trained models are converted into static rules. Only these static decision-rules are submitted for validation. Moreover, a second advantage is that the decision-rules are then implemented into the production environment independently of the development environment. It means that the Xtractree decision-rules are currently used for production across different architectures, platforms and software such as as SAS, Spark, Databricks, GCP or AWS. As previously mentioned, we are now able to reduce the time to delivery of data-driven projects with ML solutions up to 50\% thanks to the acceleration of the validation time while keeping the strengths and the innovation of AI.

\section{Experiments} \label{sec::exp}
In this section, we describe the performance of XtracTree on a real-world data set. We demonstrate, based on a trained classifier such as a RF or a decision tree, how XtracTree can simultaneously: (i) convert a classifier into a simple set of static ``if-then" rules, (ii) use these static rules to perform predictions, and (iii) describe the decision process of the classifier for a client in a simple manner, understandable by a non-expert.
\\

\begin{figure*}[t]
\centering
    \frame{\includegraphics[width=145px, height=95px]{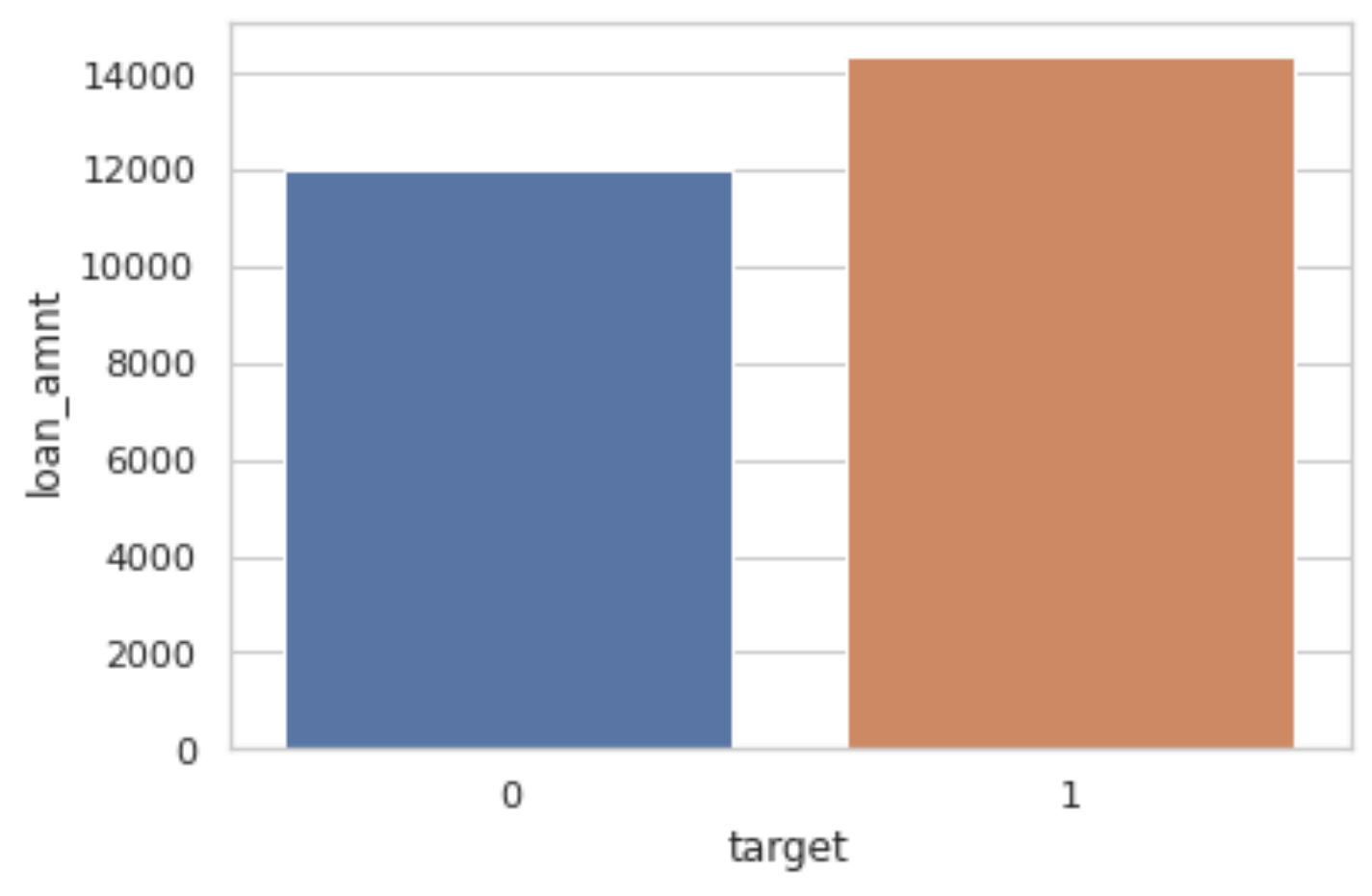}} 
    \hspace{.3cm}
    \frame{\includegraphics[width=145px, height=95px]{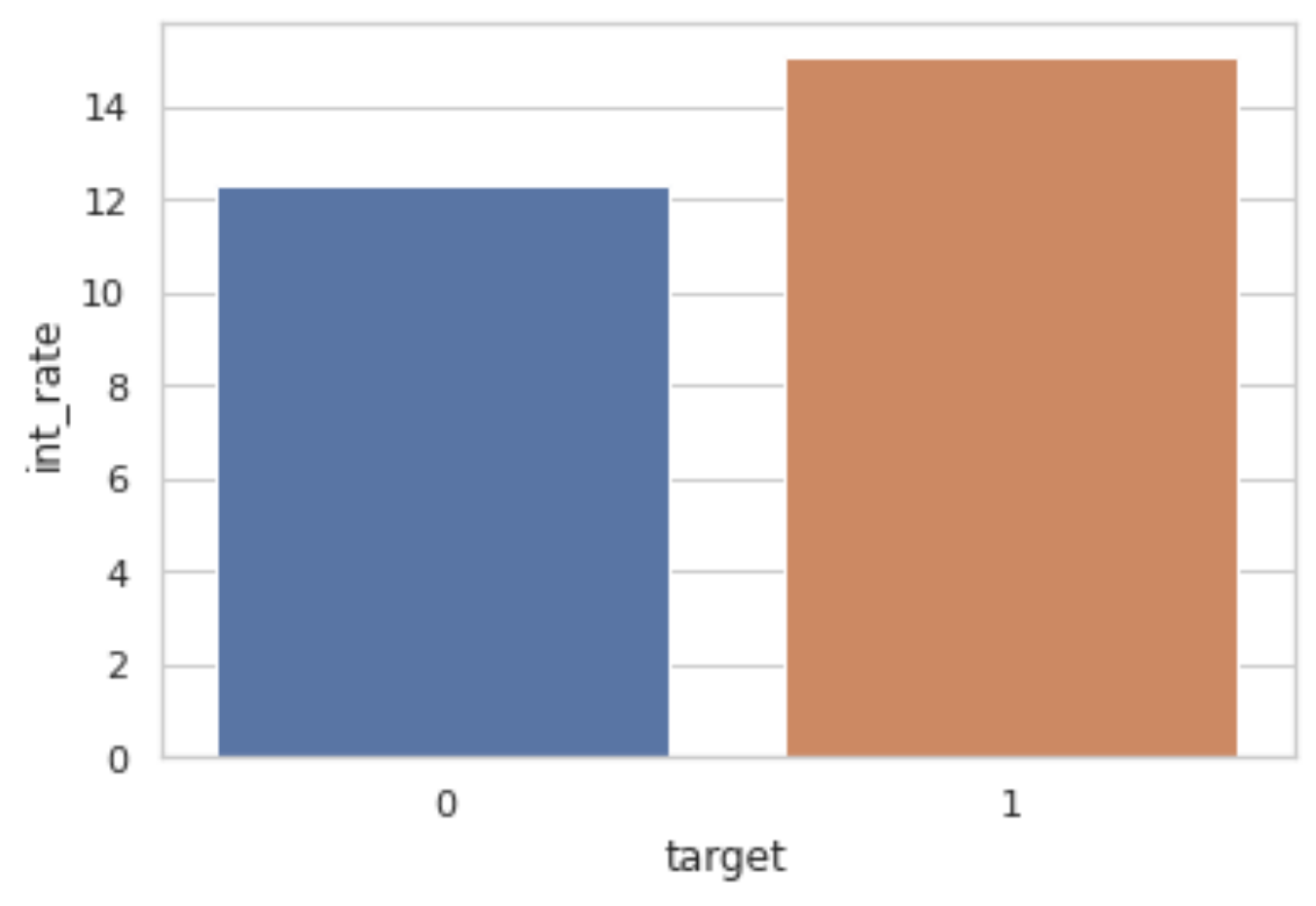}} 
    
    \vspace{.3cm}
    \frame{\includegraphics[width=145px, height=95px]{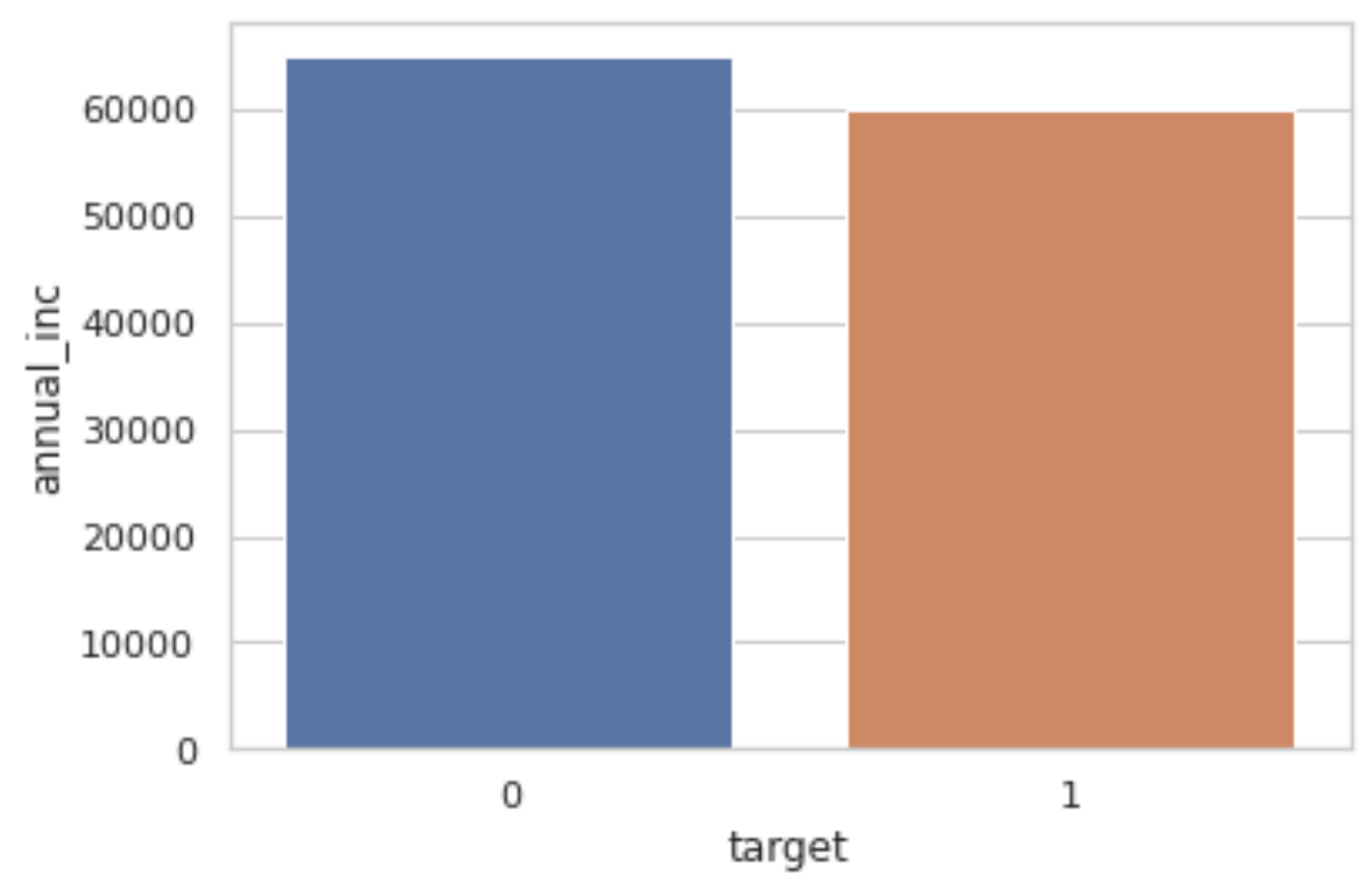}}
    \hspace{.3cm}
    \frame{\includegraphics[width=145px, height=95px]{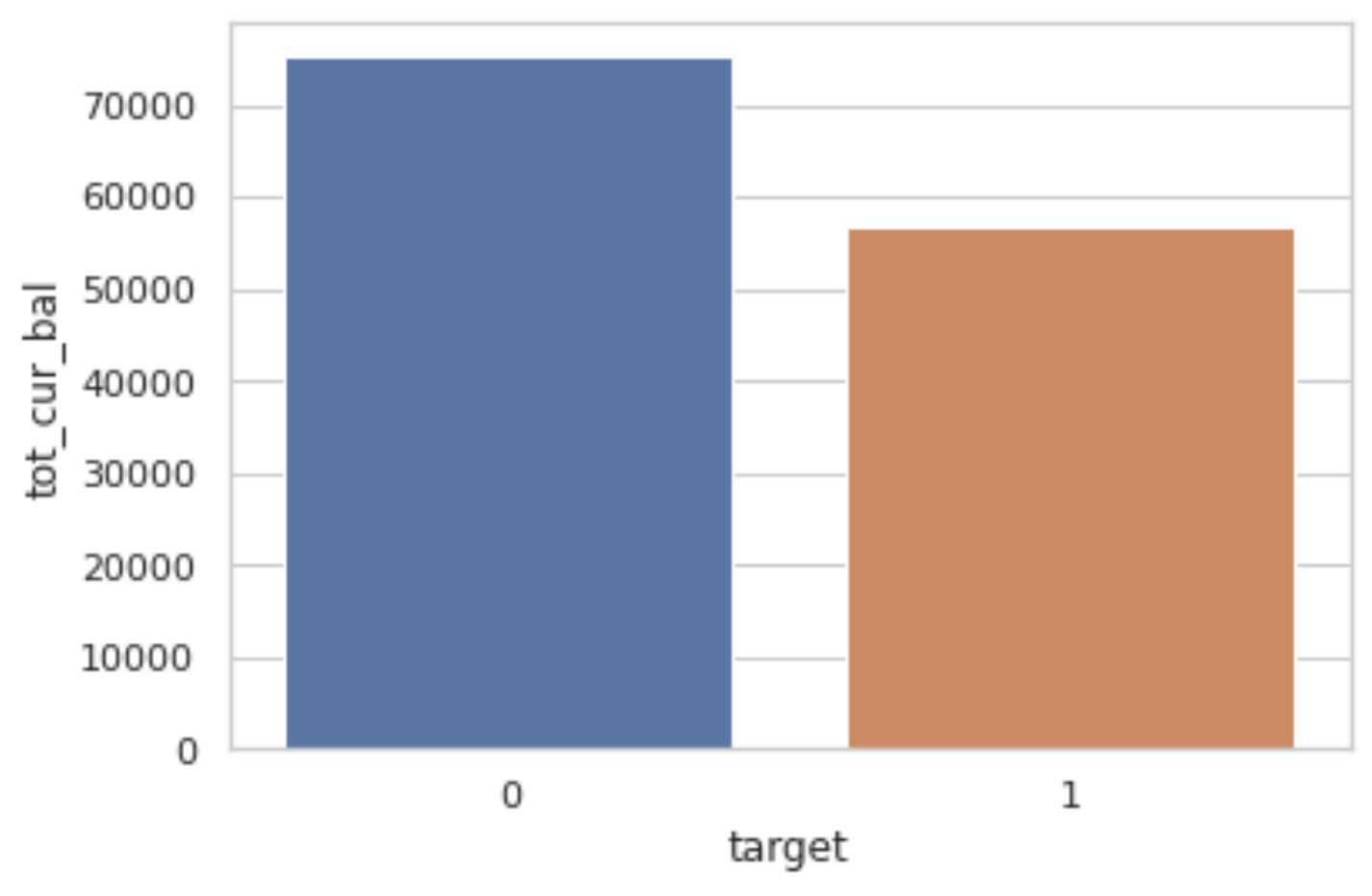}}
\caption{The target feature is labelled as 1 for defaulted loans and as 0 for fully paid loans. From the key features presented here, we notice that the defaulted loans have higher loan amounts, higher interest rates, lower annual income, and lower balance of all accounts. The loans that did default were initially financially riskier.} 
\label{fig::barplots}
\end{figure*}

\textbf{Data Availability and Data Description}
We train a bagging classifier and apply XtracTree on one real-world open data set: the lending club loan data set from the Kaggle database\footnote{https://www.kaggle.com/wendykan/lending-club-loan-data} containing 890,000 loan observations and 75 features. Each loan is labeled as ``paid in full", ``current" or ``charged-off" in the feature \textit{loan\_status}. We are interested to imitate the subscription behaviour of the clients. We, therefore, only consider the ``paid in full" and ``charged-off" loans. We label the ``paid in full" loans with the value of 1 and the ``charged-off" loans with the value of 0 in the feature called \textit{target}. We provide key distribution of the main features in Figure \ref{fig::barplots}. We divide the data into a train and a test sets with 70\% of the samples being used to build the train set and 30\% for the test set. This data set is particularly interesting because it reflects well the data sets used in retail banking activities, while being free of any confidentiality issues. 
Because of our confidentiality and compliance policies, we cannot analyze an anonymized data set containing our banking data. The risk of having information leaks including sensitive information regarding our clients is simply too high.
\\

\begin{figure}[t]
\centering
    \frame{\includegraphics[scale=0.45]{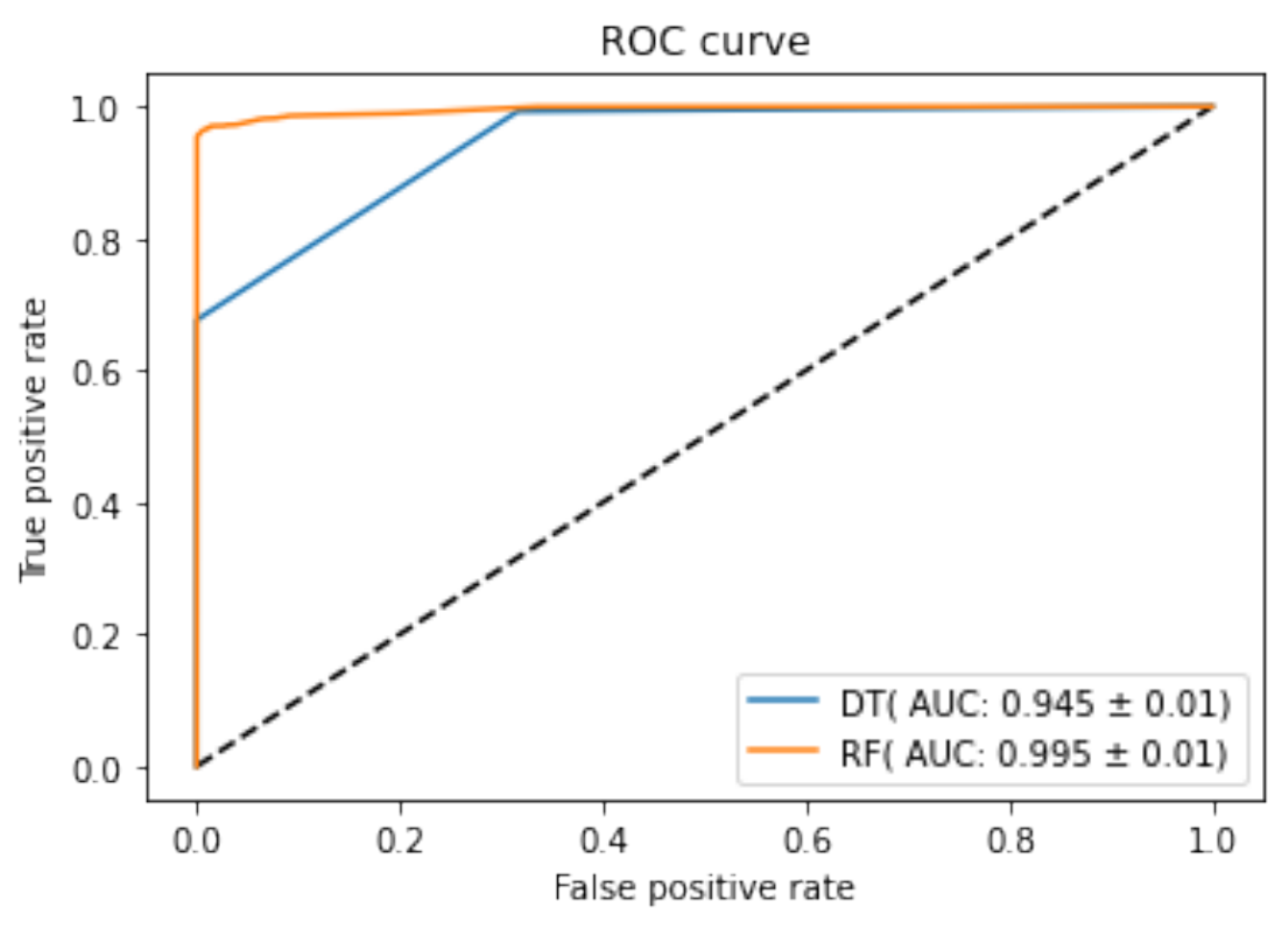}} 
\caption{We present the ROC curves for the DT and RF classifiers, as XtracTree relies on a classifier. We selected hyper-parameters allowing intuitive comprehension. The performance of the classifiers is not the core of our contribution. Bigger AUC scores can be obtained with further hyper-parameter tuning such as the depth of the classifier. A bigger classifier depth leads to a deeper XtracTree explanation and, therefore, might be less understandable at first glance for a non-expert.} 
\label{fig::roc}
\end{figure}

\textbf{Experimental Setup and Code Availability}
In our experiments, we first trained a Decision Tree (DT) classifier and then a RF classifier. Both classifiers were implemented using the scikit-learn public library \cite{scikit-learn}. We applied XtracTree with two classifiers to underline the versatility of our approach. We trained the DT with the Gini impurity as the function to measure the quality of the split \cite{scikit-learn}, maximum depth equal to 5, maximum number of features equal to 50, and maximum number of leaves equal to 10, following a grid search with cross validation. We trained the RF with the number of estimators equal to 5, maximum depth equal to 10, maximum number of features equal to 4, and the Gini impurity as the split function \cite{scikit-learn}. The parameters were optimized using a grid search with cross validation. We describe the classification performance of the two classifiers using a ROC curve (see Figure \ref{fig::roc}). Thus, we reach the core of our contribution. Here, 
We carried out the \textit{model\_2rules} function of XtracTree to build the decision rules and the \textit{predict} function to perform the predictions on the trained DT and RF classifiers. The simulations were performed on a PC computer with 16GB of RAM, Intel i7 CPU and a Tesla K80 GPU accelerator. To ensure the reproducibility of the experiments, our source code was made available at the below URL address\footnote{https://github.com/dagrate/xtractree}. 
\\

\textbf{Results and Discussions}
We structure our discussion to highlight different contributions of XtracTree. We first present the set of ``if-then" rules extracted by XtracTree. We, then, compare the probabilities computed based on the extracted set of rules of XtracTree and those computed using the classifiers. We, finally, highlight the computation of the decision path by XtracTree, which is a crucial step for the requirement of model validation. \\

In our first experiment, we present the set of ``if-then" rules extracted by XtracTree from a bagging classifier, either with a DT or a RF. We recall that in a banking environment, focusing on retail banking activities, the data is usually well structured. Bagging classifiers, therefore, are often used concurrently to boosting trees as these techniques offer high accuracy on financial data. This first experiment answers the need of the bank's departments that do not necessarily have the expertise in artificial intelligence. The rules can be converted in any language used by the business line. For instance, an Excel file can be used to implement the rules for specific projects. In Figures \ref{fig::DT} and \ref{fig::RF}, we highlight the rules extracted by XtracTree for the DT and RF classifiers. The obtained sets of rules are slightly different for the two classifiers, but some features do appear in both of them, e.g. the maturity of the loan. The set of rules allows one to describe different split decisions. One of the main contributions of XtracTree, here, is that a non-machine learning expert can easily understand the decision rules extracted by the classifier, without any need of learning how exactly these rules were computed. 
\\

\begin{figure}[t]
    \centering
    \scriptsize
    \frame{
\begin{lstlisting}[language=Python]
def estimator_tree(x, num_tree):
    if num_tree == 0:
        state = 0
        if state == 0: state = (1 if x['recoveries'] <= 0.005 
                                  else 2)
        if state == 1: state = (3 if x['last_pymnt_amnt'] <= 1214.600 
                                  else 4)
        if state == 2: return 1.0
        if state == 3: state = (5 if x['all_util'] <= 0.5 
                                  else 6)
        if state == 4: return 0.003
        if state == 5: state = (7 if x['term'] <= 48.0 
                                  else 8)
        if state == 6: state = (9 if x['total_pymnt'] <= 10010.350 
                                  else 10)
        if state == 7: state = (15 if x['total_rec_prncp'] <= 999.930 
                                   else 16)
        if state == 8: state = (17 if x['last_pymnt_amnt'] <= 223.820 
                                   else 18)
        if state == 9: state = (11 if x['funded_amnt'] <= 9012.5 
                                   else 12)
        if state == 10: state = (13 if x['total_rec_prncp'] <= 8999.730 
                                    else 14)
        if state == 11: return 0.311
        if state == 12: return 0.996
        if state == 13: return 0.904
        if state == 14: return 0.096
        if state == 15: return 0.994
        if state == 16: return 0.060
        if state == 17: return 0.086
        if state == 18: return 0.559

\end{lstlisting}}
    \caption{Set of ``if-then" rules extracted by XtracTree for the DT classifier.} 
    \label{fig::DT}
\end{figure}

\begin{figure}[t]
    \centering
    \scriptsize
    \frame{
\begin{lstlisting}[language=Python]
def estimator_tree(x, num_tree):
    if num_tree == 0:
        state = 0
        if state == 0: state = (1 if x['num_rev_tl_bal_gt_0'] <= 5.5 
                                  else 4)
        if state == 1: state = (2 if x['num_tl_op_past_12m'] <= 1.5 
                                  else 3)
        if state == 2: return 0.052
        if state == 3: return 0.069
        if state == 4: state = (5 if x['collection_recovery_fee'] \
                                  <= 0.156 else 6)
        if state == 5: return 0.032
        if state == 6: return 0.333
    elif num_tree == 1:
        state = 0
        if state == 0: state = (1 if x['num_rev_accts'] <= 0.5 else 4)
        if state == 1: state = (2 if x['term'] <= 48.0 else 3)
        if state == 2: return 0.042
        if state == 3: return 0.083
        if state == 4: state = (5 if x['installment'] <= 251.600 
                                  else 6)
        if state == 5: return 0.053
        if state == 6: return 0.073
    elif num_tree == 2:
        state = 0
        if state == 0: state = (1 if x['bc_open_to_buy'] <= 9012.5 
                                  else 4)
        if state == 1: state = (2 if x['mo_sin_rcnt_tl'] <= 9.5 
                                  else 3)
        if state == 2: return 0.078
        if state == 3: return 0.062
        if state == 4: state = (5 if x['term'] <= 48.0 else 6)
        if state == 5: return 0.040
        if state == 6: return 0.089

\end{lstlisting}}
    \caption{Set of ``if-then" rules extracted by XtracTree for the RF classifier.}
    \label{fig::RF}
\end{figure}

In our second experiment, we highlight the second contribution of XtracTree. We use the set of ``if-then" rules generated by XtracTree to predict the risk of default on a loan. In other words, we predict if a new client will most likely default on his loan based on his financial attributes. We compare the XtracTree probabilities with the probabilities computed directly by the bagging classifier. In Tables \ref{tab::dtProb} and \ref{tab::rfProb}, these probabilities are summarized for 6 arbitrarily chosen clients. Observing the results in both tables, we can notice that the predicted probabilities of XtracTree and the corresponding classifier are identical. Thus, we cannot observe any difference between XtracTree and the classifiers. In Table \ref{tab::stats}, we report some statistics, such as the mean or the quantiles, regarding the XtracTree probabilities and those of the classifiers. These results suggest that the predicted probabilities between XtracTree and the classifiers are identical. The XtracTree probabilities perfectly replicate the probabilities of the classifiers. We should remind that the probabilities of the DT and the RF are not equal, as the ROC curve and the AUC score obtained for them were different. 
Thus, Xtractree is capable of replicating perfectly the predicted probabilities of a bagging classifier without a machine learning algorithm implemented. This represents a highly versatile approach in which the ``if-then" rules can be converted into any environment or language. XtracTree can, for instance, be used to implement ``if-then" rules and build prediction probabilities with an Excel file that the line of business can use autonomously. This is not the case of a scikit-learn classifier implementation that requires specific knowledge from a machine learning expert.
\\

Our final experiment targets real-life situations from retail banking activities. A client may, for instance, want to inquiry why his application for a loan, or a credit card, has been declined. The line of business has to provide a clear set of decisions rules accompanying the decision process that led to the declined application. The process has to be understandable by non machine learning experts. We explain how XtracTree is capable of transforming the decision process of bagging classifiers leading to binary classification. In Figures \ref{fig::decisionRulesDT} and \ref{fig::decisionRulesRF}, we present the rules for the DT and  RF classifiers computed by XtracTree for evaluating the decision. XtracTree is capable to extract the splitting criteria of the classifiers used to make predictions. It then translates these criteria into a list of decisions. In the case of a RF classifier, the rules are extracted for each collection of trees, and then aggregated based on feature importance. For each decision, the feature used is described with the respective value of the client. The client value is then compared to the threshold value, inherited from the splitting criteria. The aggregation of all decisions allows us to build a comprehensive answer to the clients who may want to inquiry the decision process of their applications in front of the business line. It also provides a clear understanding of how the model works for a banking application, in the context of internal model validation and regulator validation. This feature of XtracTree helps significantly the model comprehension and validation from the business point of view.

\begin{figure}[b!]
    \centering
    \footnotesize
    \lstset{language={}}
    \begin{lstlisting}[frame=single] 
decision 1: Recovery fee under collection (=0.0) <= 0.005
decision 2: Last total payment amount received (=235.42)
            <= 1214.175
decision 3: Balance to credit limit on all trades (=92.0)
            > 0.5
decision 4: Payments received to date for total amount
            funded (=2349.84) <= 9975.0903
decision 5: The total amount committed to that loan at
            that point in time (=7200) <= 9012.5
    \end{lstlisting}
    \caption{Set of decision rules extracted by XtracTree for the DT classifier. The decisions are presented to be used by the business line. The BL can provide simple answers to the clients about the decision process carried out by the machine learning algorithm. This is a regulatory requirement.}
    \label{fig::decisionRulesDT}
\end{figure}

\begin{figure}[t!]
    \centering
    \footnotesize
    \lstset{language={}}
    \begin{lstlisting}[frame=single] 
decision 0: Number of revolving trades with 
            positive balance (=7.0) > 5.5
decision 1: Recovery fee under collection (=0.0) <= 0.1558
decision 2: Number of revolving accounts (=19.0) > 0.5
decision 3: The monthly payment owed by the borrower
            if the loan originates (=496.9) > 251.6
decision 4: Total open to buy on revolving bankcards
            (=5703.0) <= 9012.5
decision 5: Months since most recent account opened
            (=13.0) > 9.5
    \end{lstlisting}
    \caption{Set of decision rules extracted by XtracTree for the RF classifier. The decisions are presented to be used by the line of business and to provide simple answers to the clients about the decision process carried out by the machine learning algorithm.}
    \label{fig::decisionRulesRF}
\end{figure}

\begin{table}[t]
 \centering
 \caption{Probabilities computed by XtracTree and the DT classifier for arbitrary samples. The extracted set of ``if-then" rules generated by XtracTree replicates the probabilities of the DT classifier.}
 \label{tab::dtProb}
 \vspace{0.25cm}
 \small
 \begin{tabular}{cccc}
  \toprule
  Client \quad & \quad XtracTree Prob. \quad & \quad DT Classifier & \quad Difference (\%) \\
  \midrule
  0 & 0.003 & 0.003 & \textbf{0.000} \\
  1 & 1.000 & 1.000 & \textbf{0.000} \\
  2 & 0.559 & 0.559 & \textbf{0.000} \\
  3 & 0.003 & 0.003 & \textbf{0.000} \\
  4 & 0.003 & 0.003 & \textbf{0.000} \\
  5 & 0.003 & 0.003 & \textbf{0.000} \\
  \bottomrule
 \end{tabular}
\end{table}

\begin{table}[t]
 \centering
 \caption{Probabilities computed by XtracTree and the RF classifier for arbitrary samples. The extracted set of ``if-then" rules generated by XtracTree replicates the probabilities of the RF classifier.}
 \label{tab::rfProb}
 \vspace{0.25cm}
 \small
 \begin{tabular}{cccc}
  \toprule
  Client \quad & \quad XtracTree Prob. \quad & \quad RF Classifier & \quad Difference (\%) \\
  \midrule
    0 & 0.029 & 0.029 & \textbf{0.000} \\
    1 & 0.032 & 0.032 & \textbf{0.000} \\
    2 & 1.000 & 1.000 & \textbf{0.000} \\
    3 & 0.068 & 0.068 & \textbf{0.000} \\
    4 & 0.009 & 0.009 & \textbf{0.000} \\
    5 & 1.000 & 1.000 & \textbf{0.000} \\
  \bottomrule
 \end{tabular}
\end{table}

\begin{table*}[t]
    \centering
    \caption{Probability statistics for the set of ``if-then" rules extracted by XtracTree, DT and RF. The statistics indicate that the set of ``if-then" rules of XtracTree perfectly replicates the probabilities of the original classifier.}
    \label{tab::stats}
    \vspace{0.25cm}
    \small
    \begin{tabular}{ccccc}
    \toprule
    Classifier Type \quad & \quad Statistics \quad & \quad XtracTree Prob. \quad & \quad Classifier Prob. & \quad Difference (\%) \\
    \midrule
    DT & Mean & 0.003 & 0.003 & \textbf{0.000} \\
    DT & Stand. Dev. & 0.363 & 0.363 & \textbf{0.000} \\
    DT & Minimum  & 0.003 & 0.003 & \textbf{0.000} \\
    DT & 25\% Quantile & 0.003 & 0.003 & \textbf{0.000} \\
    DT & 50\% Quantile & 0.003 & 0.003 & \textbf{0.000} \\
    DT & 75\% Quantile  & 0.096 & 0.096 & \textbf{0.000} \\
    DT & Maximum & 1.000 & 1.000 & \textbf{0.000} \\
    RF & Mean & 0.003 & 0.003 & \textbf{0.000} \\
    RF & Stand. Dev. & 0.363 & 0.363 & \textbf{0.000} \\
    RF & Minimum  & 0.003 & 0.003 & \textbf{0.000} \\
    RF & 25\% Quantile & 0.003 & 0.003 & \textbf{0.000} \\
    RF & 50\% Quantile & 0.003 & 0.003 & \textbf{0.000} \\
    RF & 75\% Quantile  & 0.096 & 0.096 & \textbf{0.000} \\
    RF & Maximum & 1.000 & 1.000 & \textbf{0.000} \\
    \bottomrule
    \end{tabular}
\end{table*}

\section{Conclusion} \label{sec::ccl}
In this paper, we introduced the XtracTree algorithm aiming at addressing the issue of internal validation and regulator validation of the results provided by different ML methods in the field of retail banking. 
XtracTree is a new method capable of converting an ML bagging classifier into a set of ``if-then" rules for the requirements of the model validation, avoiding the use of  ``black-box'' ML models. 
The validation procedure emphasizes primarily the explainability of the results of ML methods used in retail banking applications. In our experiments, we showed that XtracTree is capable of using the set of ``if-then" rules to achieve the same prediction accuracy as the original DT and RF classifiers. XtracTree allows one to represent the results of ML classifiers as a set of ``if-then" rules that can be transferred to business line. The business line team can then convert these rules into the format they prefer, depending on their constraints, to deploy the model into a production environment. Moreover, we   demonstrated that XtracTree can display the splitting decisions of ML classifiers as a set of business decision rules. It provides clear business decisions for the clients, and thus increases further understanding of the model from the client's point of view. Finally, the use of Xtractree in our banking institution allowed us to increase up to 50\% the time to delivery of the proposed ML solutions. Our future work 
will focus on agnostic feature selection for transactions data in order to increase the velocity of data pre-processing and, thus, to reduce the time to solution delivery. 

\bibliographystyle{ACM-Reference-Format}
\bibliography{zc_biblio}

\end{document}